\pdfoutput=1

\documentclass[11pt,a4paper]{article}
\usepackage[hyperref]{acl2021}
\usepackage{times}
\usepackage{latexsym}

\usepackage{microtype}
\usepackage{booktabs,graphicx} 
\usepackage{float} 
\usepackage{adjustbox} 
\usepackage{multirow} 
\usepackage{tabularx} 
\usepackage{textcomp} 

\usepackage[hyphenbreaks]{breakurl}
\usepackage{breakurl}
\usepackage{url}

\aclfinalcopy 
\def\aclpaperid{2105} 

\title{Learning from the Worst: Dynamically Generated Datasets\\to Improve Online Hate Detection}

\author{Bertie Vidgen$^\dagger$, Tristan Thrush$^\ddagger$, Zeerak Waseem$^\star$, Douwe Kiela$^\ddagger$\\
  $^\dagger$The Alan Turing Institute; $^\star$University of Sheffield; $^\ddagger$Facebook AI Research\\
  \texttt{bvidgen@turing.ac.uk}}

\date{}

\begin{document}
\maketitle
\begin{abstract}
We present a human-and-model-in-the-loop process for dynamically generating datasets and training better performing and more robust hate detection models.
We provide a new dataset of ${\sim}40,000$ entries, generated and labelled by trained annotators over four rounds of dynamic data creation. It includes ${\sim}15,000$ challenging perturbations and each hateful entry has fine-grained labels for the type and target of hate. Hateful entries make up 54\% of the dataset, which is substantially higher than comparable datasets.
We show that model performance is substantially improved using this approach. Models trained on later rounds of data collection perform better on test sets and are harder for annotators to trick. They also perform better on \textsc{HateCheck}, a suite of functional tests for online hate detection.
We provide the code, dataset and annotation guidelines for other researchers to use.\footnote{Accepted at ACL 2021.}

\end{abstract}



\section{Introduction}
Accurate detection of online hate speech is important for ensuring that such content can be found and tackled scalably, minimizing the risk that harm will be inflicted on victims and making online spaces more accessible and safe.
However, detecting online hate has proven remarkably difficult and concerns have been raised about the performance, robustness, generalisability and fairness of even state-of-the-art models~\cite{Waseem2018, vidgen_challenges_2019, Caselli2020, Mishra2019,Poletto2020}.
To address these challenges, we present a human-and-model-in-the-loop process for collecting data and training hate detection models.

Our approach encompasses four rounds of data generation and model training. 
We first trained a classification model using previously released hate speech datasets. We then tasked annotators with presenting content that would trick the model and yield \textit{misclassifications}.
At the end of the round we trained a new model using the newly presented data. In the next round the process was repeated with the new model in the loop for the annotators to trick.
We had four rounds but this approach could, in principle, be continued indefinitely.

Round $1$ contains original content created synthetically by annotators.
Rounds $2$, $3$ and $4$ are split into half original content and half perturbations. 
The perturbations are challenging `contrast sets', which manipulate the original text just enough to flip the label (e.g. from `Hate' to `Not Hate')~\citep{kaushik2019learning, Gardner2020}.
In Rounds $3$ and $4$ we also tasked annotators with exploring specific types of hate and taking close inspiration from real-world hate sites to make content as adversarial, realistic, and varied as possible.

Models have lower accuracy when evaluated on test sets from later rounds as the content becomes more adversarial. Similarly, the rate at which annotators trick models also decreases as rounds progress (see Table~\ref{tab:vmer}).
At the same time, models trained on data from later rounds achieve higher accuracy, indicating that their performance improves (see Table~\ref{tab:performance}).
We verify improved model performance by evaluating them against the \textsc{HateCheck} functional tests \cite{rottger2020hatecheck}, with accuracy improving from $60$\% in Round $1$ to $95$\% in Round $4$.
In this way the models `learn from the worst' because as the rounds progress (a) they become increasingly accurate in detecting hate which means that (b) annotators have to provide more challenging content in order to trick them.

We make three contributions to online hate classification research.
First, we present a human-and-model-in-the-loop process for training online hate detection models.
Second, we present a dataset of $40,000$ entries, of which $54$\% are hate. It includes fine-grained annotations by trained annotators for label, type and target (where applicable). 
Third, we present high quality and robust hate detection models.
All data, code and annotation guidelines are available.\footnote{\url{https://github.com/bvidgen/Dynamically-Generated-Hate-Speech-Dataset}}

\section{Background}
\label{sec:lit}

\paragraph{Benchmark datasets}
Several benchmark datasets have been put forward for online hate classification~\cite{Waseem2016, waseem-2016-racist, Davidson2017, Founta2018a, Mandl2019, Zampieri2019, Zampieri2020a, Vidgen2020b, vidgen-etal-2021-introducing}.
These datasets offer a comparison point for detection systems and have focused the field's attention on important subtasks, such as classification across different languages, domains and targets of hate. 
Performance on some benchmark datasets has increased substantially through the use of more advanced models. For instance, in the original \newcite{Waseem2016} paper in $2016$, the authors achieved an F1 of $0.74$. By $2018$ this had increased to $0.93$ \cite{Pitsilis2018}. 

Numerous problems have been identified with hate speech training datasets, such as lacking linguistic variety, being inexpertly annotated and degrading over time \cite{vidgen_challenges_2019, Poletto2020}.
\citet{Vidgen2020} examined $63$ open-source abusive language datasets and found that $27$ ($43\%$) were sourced from Twitter \cite{Vidgen2020}. In addition, many datasets are formed with bootstrapped sampling, such as keyword searches, due to the low prevalence of hate speech `in the wild' \cite{Vidgen2019b}. Such bootstrapping can substantially bias the nature and coverage of datasets \cite{Wiegand2019}. Models trained on historical data may also not be effective for present-day hate classification models given how quickly online conversations evolve \cite{Nobata2016}. 

\paragraph{Model limitations}
Systems trained on existing datasets have been shown to lack accuracy, robustness and generalisability, creating a range of false positives and false negatives \cite{schmidtSurveyHateSpeech2017a,Mishra2019, Vidgen2020, rottger2020hatecheck, mathew2020hatexplain}. These errors often make models unsuitable for use in downstream tasks, such as moderating online content or measuring online hate.

False positives are non-hateful entries which are incorrectly classified as hateful.
\newcite{Vidgen2020b} report that $29$\% of errors from a classifier for East Asian prejudice are due to lexical similarities between hateful and non-hateful entries, such as abuse directed towards out-of-scope targets being misclassified as Sinophobic.
Other research shows that some identity terms (e.g. `gay') are substantially more likely to appear in toxic content in training datasets, leading models to overfit on them \cite{Dixon2018,Kennedy2020}.
Similarly, many models overfit on the use of slurs and pejorative terms, treating them as hateful irrespective of how they are used \cite{Waseem2018,Davidson2017,Kurrek2020,Palmer2020}. This is problematic when the terms are used as part of counter speech \cite{Wright2017,Chung2019} or have been reclaimed by the targeted group \cite{Waseem2018, Sap2019}.
Models can also misclassify interpersonal abuse and incivil language as hateful \cite{wulcz2017, Zampieri2019,  Palmer2020}. 

False negatives are hateful entries which are incorrectly classified as non-hateful. 
\newcite{Grondahl2018} show that making simple changes such as inserting spelling errors, using leetspeak\footnote{Leetspeak refers to the obfuscation of words by replacing letters with similar looking numbers and symbols.}, changing word boundaries, and appending words can lead to misclassifications of hate. \citet{Hosseini2017} also investigate how detection models can be attacked and report similar findings. In other cases, false negatives can be provoked by changing the `sensitive' attribute of hateful content, such as changing the target from `gay' to `black' people \cite{Garg2019}. This can happen when models are trained on data which only contains hate directed against a limited set of targets \cite{Salminen2020}.
Another source of false negatives is when classification systems are applied to out-of-domain settings, such as system trained on Twitter data being applied to data from Gab \cite{Karan2018, Pamungkas2020, Swamy2019, Basile2019, Salminen2020}.
Subtle and implicit forms of hate speech can also create false negatives \cite{Vidgen2018, Palmer2020, mathew2020hatexplain}, as well as more `complex' forms of speech such as sarcasm, irony, adjective nominalization and rhetorical questions \cite{Caselli2020, vidgen_challenges_2019}. 

\paragraph{Dynamic benchmarking and contrast sets}
Addressing the numerous flaws of hate detection models is a difficult task. The problem may partly lie in the use of static benchmark datasets and fixed model evaluations. In other areas of Natural Language Processing, several alternative model training and dataset construction paradigms have been presented, involving dynamic and iterative approaches.
In a dynamic dataset creation setup, annotators are incentivised to produce high-quality `adversarial' samples which are challenging for baseline models, repeating the process over multiple rounds \cite{nie-etal-2020-adversarial}. This offers a more targeted way of collecting data.
\newcite{Dinan2019} ask crowd-workers to `break' a BERT model trained to identify toxic comments and then retrain it using the new examples. Their final model is more robust to complex forms of offensive content, such as entries with figurative language and without profanities.

Another way of addressing the limitations of static datasets is through creating `contrast sets' of perturbations \cite{kaushik2019learning, Gardner2020}. By making minimal label-changing modifications that preserve `lexical/syntactic artifacts present in the original example' \cite[p. 1308]{Gardner2020} the risk of overfitting on spurious correlations is minimized.
Perturbations have only received limited attention in the context of hate detection. \newcite{Samory2020} create $2,000$ `hard-to-classify' not-sexist examples which contrast sexist examples in their dataset. They show that fine-tuning a BERT model with the contrast set produces more robust classification system.

Dynamic benchmarking and contrast sets highlight the effectiveness of developing datasets in a directed and adaptive way, ensuring that models learn from and are evaluated on the most challenging content. However, to date, these approaches remain under-explored for hate speech detection and to the best of our knowledge no prior work in hate speech detection has combined the two approaches within one system.

\section{Dataset labels}
Previous research shows the limitations of using only a binary labelling schema (i.e., `Hate' and `Not Hate').
However, there are few established taxonomies and standards in online hate research, and most of the existing datasets have been labelled with very different schemas. The hierarchical taxonomy we present aims for a balance between granularity versus conceptual distinctiveness and annotation simplicity, following the guidance of \newcite{Nickerson2013}.
All entries are assigned to either `Hate' or `Not Hate'. `Hate' is defined as ``abusive speech targeting specific group characteristics, such as ethnic origin, religion, gender, or sexual orientation.''~\cite{Warner2012}.
For `Hate', we also annotate secondary labels for the type and target of hate. The taxonomy for the type of hate draws on and extends previous work, including \newcite{Waseem2016, vidgen_challenges_2019, Zampieri2019}.

\subsection{Types of hate}
\paragraph{Derogation} Content which explicitly attacks, demonizes, demeans or insults a group.
This resembles similar definitions from \newcite{Davidson2017}, who define hate as content that is `derogatory', \newcite{Waseem2016} who include `attacks' in their definition, and \newcite{Zampieri2019} who include `insults'.

\paragraph{Animosity} Content which expresses abuse against a group in an implicit or subtle manner. It is similar to the `implicit' and `covert' categories used in other taxonomies \cite{Waseem2017, Vidgen2018, Kumar2018a}.

\paragraph{Threatening language} Content which expresses intention to, support for, or encourages inflicting harm on a group, or identified members of the group. 
This category is used in datasets by \newcite{Hammer2014}, \newcite{Golbeck2017} and \newcite{Anzovino2018}. 

\paragraph{Support for hateful entities} Content which explicitly glorifies, justifies or supports hateful actions, events, organizations, tropes and individuals~(collectively, `entities’). 

\paragraph{Dehumanization} Content which `perceiv[es] or treat[s] people as less than human' \cite{Haslam2016}. It often involves describing groups as leeches, cockroaches, insects, germs or rats \cite{Mendelsohn2020}. 

\subsection{Targets of hate}
Hate can be targeted against any vulnerable, marginalized or discriminated-against group. We provided annotators with a non-exhaustive list of 29 identities to focus on (e.g., women, black people, Muslims, Jewish people and gay people), as well as a small number of intersectional variations (e.g., `Muslim women'). They are given in Appendix~\ref{sec:appendix-identities}.
Some identities were considered out-of-scope for Hate, including men, white people, and heterosexuals. 

\section{Annotation}
Data was annotated using an open-source web platform for dynamic dataset creation and model benchmarking.\footnote{\url{https://anonymized-url}} The platform supports human-and-model-in-the-loop dataset creation for a variety of NLP tasks. 
Annotation was overseen by two experts in online hate.
The annotation process is described in the following section.
Annotation guidelines were created at the start of the project and then updated after each round in response to the increased need for detail from annotators. 
We followed the guidance for protecting and monitoring annotator well-being provided by \newcite{vidgen_challenges_2019}.
20 annotators were recruited. They received extensive training and feedback during the project.
Full details on the annotation team are given in Appendix~\ref{sec:appendix-ds}.
The small pool of annotators was driven by the logistical constraints of hiring and training them to the required standard and protecting their welfare given the sensitivity and complexity of the topic. Nonetheless, it raises the potential for bias. We take steps to address this in our test set construction and provide an annotator ID with each entry in our publicly-released dataset to enable further research into this issue.


\section{Dataset formation}
The dataset was generated over four rounds, each of which involved ${\sim}10,000$ entries.
The final dataset comprises $41,255$ entries, as shown in Table~\ref{tab:data-summary}. The ten groups that are targeted most often are given in Table~\ref{tab:data-targets}. Entries could target multiple groups.
After each round, the data was split into training, dev and test splits of $80$\%, $10$\% and $10$\%, respectively. 
Approximately half of the entries in the test sets are produced by annotators who do not appear in the training and dev sets~(between $1$ and $4$ in each round). This makes the test sets more challenging and minimizes the risk of annotator bias given our relatively small pool of annotators~\cite{geva-etal-2019-modeling}. The other half of each test set consists of content from annotators who do appear in the training and dev sets.

Rounds 2, 3 and 4 contain perturbations. In 18 cases the perturbation does not flip the label. This mistake was only identified after completion of the paper and is left in the dataset. These cases can be identified by checking whether original and perturbed entries that have been linked together have the same labels (e.g., whether an original and perturbation are both assigned to 'Hate').

\begin{table*}[ht]
\centering
\begin{tabular}{l|l|l|llll}
\toprule
\textbf{Label}       & \textbf{Type}        & \textbf{Total}       & \textbf{R1}          & \textbf{R2}          & \textbf{R3}          & \textbf{R4}  \\
\toprule
Hate                 & Not given            & $7,197$                & $7,197$                & $0$                    & $0$                    & $0$    \\     
                     & Animosity            & $3,439$                & $0 $                   & $758$                  & $1,206$                & $1,475$  \\        
                     & Dehumanization       & $906$                  & $0 $                   & $261$                  & $315$                  & $330$   \\        
                     & Derogation           & $9,907$                & $0$                    & $3,574$                & $3,206$                & $3,127$  \\        
                     & Support              & $207$                  & $0$                    & $41$                   & $104$                  & $62$    \\          
                     & Threatening          & $606$                  & $0 $                   & $376$                  & $148$                  & $82$    \\
\toprule
                     & Total                & $22,262$               & $7,197$                 & $5,010$               & $4,979$              & $5,076$  \\
\toprule
Not Hate             & /                    & $18,993$                & $3,960$                 & $4,986$                 & $4,971$               & $5,076 $ \\
\toprule
All                  & TOTAL                & $41,255$                & $11,157$                & $9,996$                 & $9,950$                & $10,152$ \\
\bottomrule
\end{tabular}
\caption{\label{tab:data-summary}Summary of data collected in each round}
\end{table*}

\begin{table}[t]
\centering
\begin{tabular}{lc}
\toprule
\textbf{Target} & \textbf{Number of entries} \\
\toprule
Black people & $2,278$ \\
Women & $2,192$ \\
Jewish people & $1,293$ \\
Muslims & $1,144$ \\
Trans people & $972$ \\
Gay people & $875$ \\
Immigrants & $823$ \\
Disabled people & $575$ \\
Refugees & $533$ \\
Arabs & $410$ \\
\bottomrule
\end{tabular}
\caption{\label{tab:data-targets}Most common targets of hate in the dataset}
\end{table}


\paragraph{Target model implementation}
Every round has a model in the loop, which we call the `target model'. The target model is always trained on a combination of data collected in the previous round(s). For instance, \textbf{M2} is the target model used in \textbf{R2}, and was trained on \textbf{R1} and \textbf{R0} data.
For consistency, we use the same model architecture everywhere, specifically RoBERTa \cite{liu2019roberta} with a sequence classification head. We use the implementation from the Transformers \cite{Wolf2019HuggingFacesTS} library. More details are available in appendix \ref{sec:appendix-model}.

For each new target model, we identify the best sampling ratio of previous rounds' data using the dev sets.
\textbf{M1} is trained on \textbf{R0} data.
\textbf{M2} is trained on \textbf{R0} data and \textbf{R1} upsampled to a factor of five.
\textbf{M3} is trained on the data used for \textbf{M2} and \textbf{R2} data upsampled to a factor of one hundred.
\textbf{M4} is trained on the data used for \textbf{M3} and one lot of the \textbf{R3} data.


\subsection{Round 1 (R1)}
The target model in \textbf{R1} is \textbf{M1}, a RoBERTa model trained on \textbf{R0} which consists of $11$ English language training datasets for hate and toxicity taken from \url{hatespeechdata.com}, as reported in \newcite{Vidgen2020}. It includes widely-used datasets provided by \newcite{waseem-2016-racist}, \newcite{Davidson2017} and \newcite{Founta2018a}. It comprises $468,928$ entries, of which $22$\% are hateful/toxic.
The dataset was anonymized by replacing usernames, indicated by the `@' symbol. URLs were also replaced with a special token.
In \textbf{R1}, annotators were instructed to enter synthetic content into the model that would trick \textbf{M1} using their own creativity and by exploiting any model weaknesses they identified through the real-time feedback.

All entries were validated by one other annotator and entries marked as incorrect were sent for review by expert annotators. This happened with $1,011$ entries. 
$385$ entries were excluded for being entirely incorrect. In the other cases, the expert annotator decided the final label and/or made minor adjustments to the text.
The final dataset comprises $11,157$ entries of which $7,197$ are `Hate' (65\%) and $3,960$ are `Not Hate' ($35$\%). The type and target of Hate was not recorded by annotators in \textbf{R1}.

\subsection{Round 2 (R2)}
A total of $9,996$ entries were entered in \textbf{R2}.
The hateful entries are split between Derogation ($3,577$, $72$\%), Dehumanization ($255$, $5$\%), Threats ($380$, $8$\%), Support for hateful entities ($39$, $1$\%) and Animosity ($759$, $15$\%).
In \textbf{R2} we gave annotators adversarial `pivots' to guide their work, which we identified from a review of previous literature (see Section~\ref{sec:lit}).
The $10$ hateful and $12$ not hateful adversarial pivots, with examples and a description, are given in Appendix~\ref{sec:appendix-pivots}.
Half of \textbf{R2} comprises originally entered content and the other half comprises perturbed contrast sets.

Following \newcite{Gardner2020}, perturbations were created offline without feedback from a model-in-the-loop. Annotators were given four main points of guidance: (1) ensure perturbed entries are realistic, (2) firmly meet the criteria of the flipped label and type, (3) maximize diversity within the dataset in terms of type, target and how entries are perturbed and (4) make the least changes possible while meeting (1), (2) and (3).
Common strategies for perturbing entries included changing the target (e.g., from `black people' to `the local council'), changing the sentiment (e.g. `It's \textit{wonderful} having gay people round here'), negating an attack (e.g. `Muslims are \textit{not} a threat to the UK') and quoting or commenting on hate. 

Of the original entries, those which fooled \textbf{M1} were validated by between three and five other annotators. Every perturbation was validated by one other annotator.
Annotators could select: (1) correct if they agreed with the label and, for Hate, the type/target, (2) incorrect if the label was wrong or (3) flag if they thought the entry was unrealistic and/or they agreed with the label for hate but disagreed with the type or target.
Krippendorf's alpha is $0.815$ for all original entries if all `flagged' entries are treated as `incorrect', indicating extremely high levels of agreement \cite{Hallgren2012a}.
All of the original entries identified by at least two validators as incorrect/flagged, and perturbations which were identified by one validator as incorrect/flagged, were sent for review by an expert annotator. This happened in $760$ cases in this round.

\paragraph{Lessons from R2} The validation and review process identified some limitations of the \textbf{R2} dataset.
First, several `template' statements were entered by annotators. These are entries which have a standardized syntax and/or lexicon, with only the identity changed, such as `[Identity] are [negative attribute]'. When there are many cases of each template they are easy for the model to correctly classify because they create a simple decision boundary.
Discussion sessions showed that annotators used templates (i) to ensure coverage of different identities (an important consideration in making a \textit{generalisable} online hate classifier) and (ii) to maximally exploit model weaknesses to increase their model error rate. We banned the use of templates.
Second, in attempting to meet the `pivots' they were assigned, some annotators created unrealistic entries. We updated guidance to emphasize the importance of realism.
Third, the pool of 10 trained annotators is large for a project annotating online hate but annotator biases were still produced. 
Model performance was high in \textbf{R2} when evaluated on a training/dev/test split with all annotators stratified. We then held out some annotators' content and performance dropped substantially. We use this setup for all model evaluations.




\subsection{Round 3 (R3)}
In \textbf{R3} annotators were tasked with finding real-world hateful online content to inspire their entries. All real-world content was subject to at least one substantial adjustment prior to being presented to the model.
$9,950$ entries were entered in \textbf{R3}.
The hateful entries are split between Derogation ($3,205$, $64$\%), Dehumanization ($315$, $6$\%), Threats ($147$, $3$\%), Support for hateful entities ($104$, $2$\%) and Animosity ($1,210$, $24$\%).
Half of \textbf{R3} comprises originally entered content ($4,975$) and half comprises perturbed contrast sets ($4,975$).

The same validation procedure was used as with \textbf{R2}. Krippendorf's alpha was $0.55$ for all original entries if all `flagged' entries are treated as `incorrect', indicating moderate levels of agreement~\cite{Hallgren2012a}.
This is lower than \textbf{R2}, but still comparable with other hate speech datasets (e.g., \citet{wulczynExMachinaPersonal2017} achieve Krippnedorf's alpha of $0.45$).
Note that more content is labelled as Animosity compared with \textbf{R2} ($24$\% compared with $15$\%), which tends to have higher levels of disagreement. $981$ entries were reviewed by the expert annotators.


\subsection{Round 4 (R4)}
As with \textbf{R3}, annotators searched for real-world hateful online content to inspire their entries. In addition, each annotator was given a target identity to focus on (e.g., Muslims, women, Jewish people). 
The annotators (i) investigated hateful online forums and communities relevant to the target identity to find the most challenging and nuanced content and (ii) looked for challenging non-hate examples, such as neutral discussions of the identity.
$10,152$ entries were entered in \textbf{R4}, comprising $5,076$ `Hate' and $5,076$ `Not Hate'.
The hateful entries are split between Derogation ($3,128$, $62$\%), Dehumanization ($331$, $7$\%), Threats ($82$, $2$\%), Support for hateful entities ($61$, $1$\%) and Animosity ($1,474$, $29$\%).
Half of \textbf{R4} comprises originally entered content ($5,076$) and half comprises perturbed contrast sets ($5,076$).
The same validation procedure was used as in \textbf{R2} and \textbf{R3}. Krippendorf's alpha was $0.52$ for all original entries if all `flagged' entries are treated as `incorrect', indicating moderate levels of agreement~\cite{Hallgren2012a}.
This is similar to \textbf{R2}. $967$ entries were reviewed by the expert annotators following the validation process.


\begin{table*}[t]
    \centering
    \begin{tabular}{l|c|c|c|ccccc}\toprule
    \textbf{Round} & \textbf{Total} & \textbf{Not} & \textbf{Hate} & \small{\textbf{Animosity}} & \small{\textbf{\begin{tabular}{@{}l@{}}Dehuman\\-ization\end{tabular}}} & \small{\textbf{Derogation}} & \small{\textbf{Support}} & \small{\textbf{Threatening}} \\\midrule
    R1  & $54.7\%$ & $64.6\%$ & $49.2\%$ & -        & -        & -        & -        & - \\
    R2  & $34.3\%$ & $38.9\%$ & $29.7\%$ & $40.1\%$ & $25.5\%$ & $28.7\%$ & $53.8\%$ & $18.4\%$  \\
    R3  & $27.8\%$ & $20.5\%$ & $35.1\%$ & $53.8\%$ & $27.9\%$ & $29.2\%$ & $59.6\%$ & $17.7\%$  \\
    R4  & $27.7\%$ & $23.7\%$ & $31.7\%$ & $44.5\%$ & $21.1\%$ & $26.9\%$ & $49.2\%$ & $18.3\%$  \\
    \midrule
    All & $36.6\%$ & $35.4\%$ & $37.7\%$ & $46.4\%$ & $24.8\%$ & $28.3\%$ & $55.4\%$ & $18.2\%$  \\
\bottomrule
    \end{tabular}
    \caption{Error rate for target models in each round. Error rate decreases as the rounds progress, indicating that models become harder to trick. Annotators were not given real-time feedback on whether their entries tricked the model when creating perturbations. More information about tuning is available in appendix \ref{sec:appendix-model}}
    \label{tab:vmer}
\end{table*}

\section{Model performance}

In this section, we examine the performance of models on the collected data, both when used in-the-loop during data collection (measured by the model error rate on new content shown by annotators), as well as when separately evaluated against the test sets in each round's data. We also examine how models generalize by evaluating them on the out-of-domain suite of diagnostic functional tests in \textsc{HateCheck}.

\subsection{Model error rate}
The model error rate is the rate at which annotator-generated content tricks the model. It decreases as the rounds progress, as shown in Table~\ref{tab:vmer}.
\textbf{M1}, which was trained on a large set of public hate speech datasets, was the most easily tricked, even though many annotators were learning and had not been given advice on its weaknesses. 
$54.7\%$ of entries tricked it, including $64.6\%$ of Hate and $49.2\%$ of Not Hate.
Only $27.7\%$ of content tricked the final model (\textbf{M4}), including $23.7\%$ of Hate and $31.7\%$ of Not Hate.
The type of hate affected how frequently entries tricked the model. 
In general, more explicit and overt forms of hate had the lowest model error rates, with threatening language and dehumanization at $18.2\%$ and $24.8\%$ on average, whereas support for hateful entities and animosity had the highest error ($55.4\%$ and $46.4\%$ respectively).
The model error rate falls as the rounds progress but nonetheless this metric potentially still underestimates the increasing difficulty of the rounds and the improvement in the models. Annotators became more experienced and skilled over the annotation process, and entered progressively more adversarial content. As such the content that annotators enter becomes far harder to classify in the later rounds, which is also reflected in all models' lower performance on the later round test sets (see Table~\ref{tab:performance}).


\subsection{Test set performance}
Table~\ref{tab:performance} shows the macro F1 of models trained on different combinations of data, evaluated on the test sets from each round (see Appendix~\ref{sec:appendix-dev} for dev set performance). 
The target models achieve lower scores when evaluated on test sets from the later rounds, demonstrating that the dynamic approach to data collection leads to increasingly more challenging data. The highest scores for \textbf{R3} and \textbf{R4} data are in the mid-$70$s, compared to the high $70$s in \textbf{R2} and low $90$s in \textbf{R1}.
Generally, the target models from the later rounds have higher performance across the test sets. For instance, \textbf{M4} is the best performing model on \textbf{R1}, \textbf{R2} and \textbf{R4} data. It achieves 75.97 on the \textbf{R4} data whereas \textbf{M3} achieves 74.83 and \textbf{M2} only 60.87.
A notable exception is \textbf{M1} which outperforms \textbf{M2} on the \textbf{R3} and \textbf{R4} test sets.

Table~\ref{tab:performance} presents the results for models trained on just the training sets from each round (with no upsampling), indicated by M(RX only). In general the performance is lower than the equivalent target model. For instance, \textbf{M4} achieves macro F1 of 75.97 on the \textbf{R4} test data. \textbf{M(R3 only)} achieves 73.16 on that test set and \textbf{M(R4 only)} just 69.6. In other cases, models which are trained on just one round perform well on some rounds but are far worse on others. Overall, building models cumulatively leads to more consistent performance. Table~\ref{tab:performance} also shows models trained on the cumulative rounds of data with no upsampling, indicated by M(RX+RY). In general, performance is lower without upsampling; the F1 of \textbf{M3} is 2 points higher on the \textbf{R3} test set than the equivalent non-upsampled model (\textbf{M(R0+R1+R2)}).


\begin{table*}[t]
    \centering
    \begin{tabular}{l|cccc}\toprule
    \textbf{Model} & \textbf{R1} & \textbf{R2} & \textbf{R3} & \textbf{R4} \\\midrule
    M1 (R1 Target) & 44.84$\pm$1.1 & 54.42$\pm$0.45 & 66.07$\pm$1.03 & 60.91$\pm$0.4 \\ 
    M2 (R2 Target) & 90.17$\pm$1.42 & 66.05$\pm$0.67 & 62.89$\pm$1.26 & 60.87$\pm$1.62 \\ 
    M3 (R3 Target) & 91.37$\pm$1.26 & 77.14$\pm$1.26 & \textbf{76.97$\pm$0.49} & 74.83$\pm$0.92 \\ 
    M4 (R4 Target) & \textbf{92.01$\pm$0.6} & \textbf{78.02$\pm$0.91} & 75.89$\pm$0.62 & \underline{\textbf{75.97$\pm$0.96}} \\\midrule
    M(R1 only) & \underline{\textbf{92.20$\pm$0.55}} & 62.87$\pm$0.63 & 47.67$\pm$1.04 & 52.37$\pm$1.27\\ 
    M(R2 only) & 80.73$\pm$0.4 & 76.52$\pm$0.7 & \underline{\textbf{77.43$\pm$0.51}} & \textbf{74.88$\pm$0.85}\\ 
    M(R3 only) & 72.71$\pm$1.05 & \underline{\textbf{78.55$\pm$0.71}} & 74.14$\pm$1.5 & 73.16$\pm$0.58\\ 
    M(R4 only) & 72.26$\pm$1.3 & 76.78$\pm$1.65 & 77.21$\pm$0.43 & 69.6$\pm$0.6\\\midrule
    M(R0+R1) & 88.78$\pm$0.89 & 66.15$\pm$0.77 & 67.15$\pm$1.11 & 63.44$\pm$0.26\\ 
    M(R0+R1+R2) & 91.09$\pm$0.37 & 74.73$\pm$0.95 & 74.73$\pm$0.46 & 71.59$\pm$0.59\\ 
    M(R0+R1+R2+R3) & \textbf{91.17$\pm$0.99} & 77.03$\pm$0.72 & 74.6$\pm$0.48 & \textbf{73.94$\pm$0.94}\\ 
    M(R0+R1+R2+R3+R4) & 90.3$\pm$0.96 & \textbf{77.93$\pm$0.84} & \textbf{76.79$\pm$0.24} & 72.93$\pm$0.56\\\bottomrule
    \end{tabular}
    \caption{Macro F1 with standard deviation over 5 training rounds, evaluated on each rounds' test set. Early-stopping is performed on the latest dev set for each round where dev results are obtained at least once per epoch, out of four epochs.}
    \label{tab:performance}
\end{table*}

\subsection{HateCheck}
To better understand the weaknesses of the target models from each round, we apply them to \textsc{HateCheck}, as presented by \newcite{rottger2020hatecheck}.
\textsc{HateCheck} is a suite of functional tests for hate speech detection models, based on the testing framework introduced by \citet{ribeiro2020beyond}. It comprises $29$ tests, of which $18$ correspond to distinct expressions of hate and the other $11$ are non-hateful contrasts. The selection of functional tests is motivated by a review of previous literature and interviews with $21$ NGO workers. From the $29$ tests in the suite, $3,728$ labelled entries are generated in the dataset of which $69$\% are `Hate' and $31$\% are `Not Hate'. 

\begin{figure}[t]
    \centering
    \includegraphics[width=0.98\columnwidth]{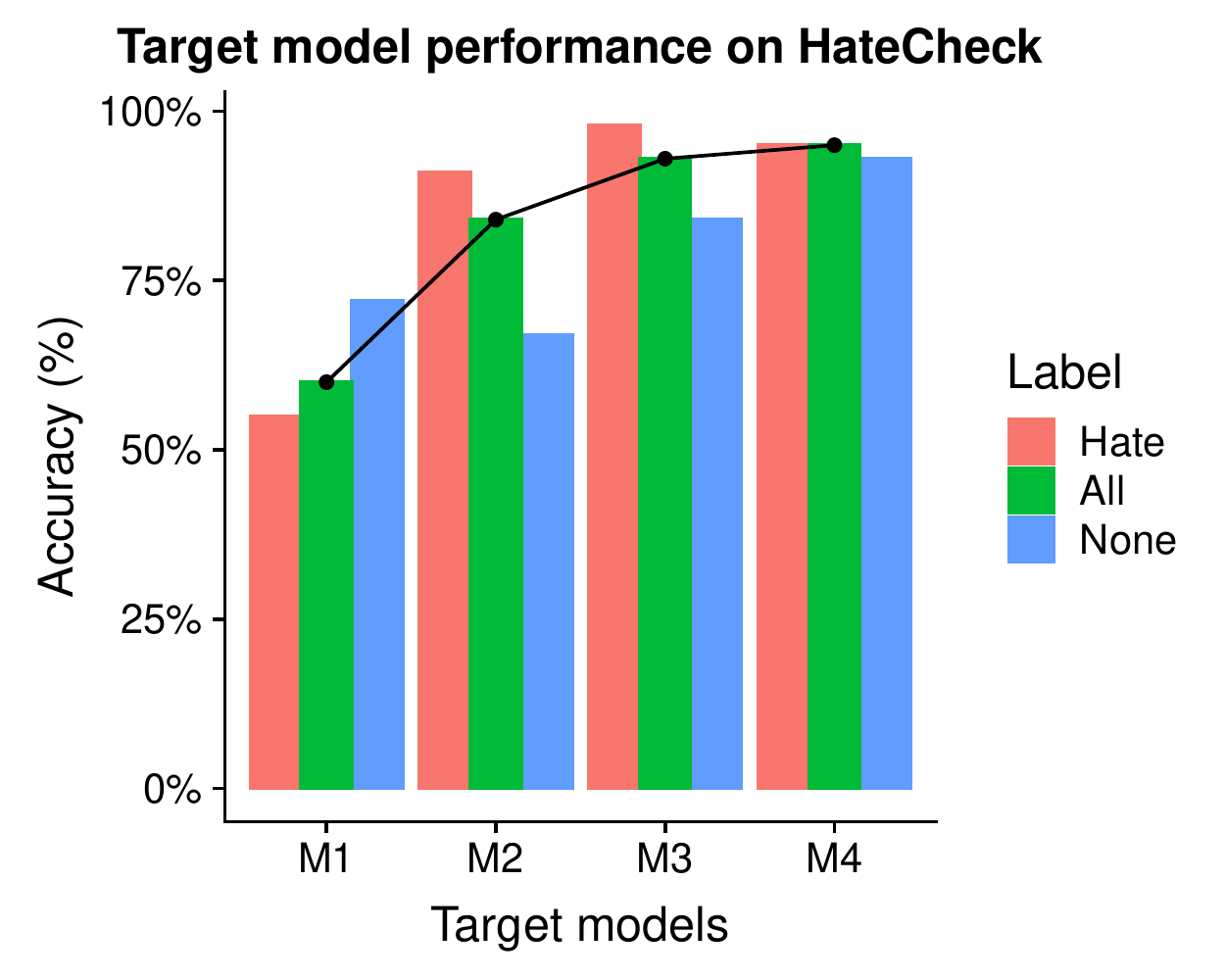}
    \caption{Performance of target models on the \textsc{HateCheck} test suite.}
    \label{fig:saturation}
\end{figure}

Performance of target models trained on later rounds is substantially higher, increasing from $60$\% (on both `Hate' and `Not Hate') combined for \textbf{M1} to $95$\% for \textbf{M4}.
Performance is better than all four models evaluated by \newcite{rottger2020hatecheck}, of which Perspective's toxicity classifier\footnote{See: \url{https://www.perspectiveapi.com/\#/home}.} is best performing with $77$\% overall accuracy, including $90$\% on `Hate' and $48$\% on `Not Hate'. 
Notably, the performance of \textbf{M4} is consistent across both `Hate' and `Not Hate', achieving $95$\% and $93$\% respectively. This is in contrast to earlier target models, such as \textbf{M2} which achieves $91$\% on `Hate' but only $67$\% on `Not Hate' (note that this is actually a \textit{reduction} in performance from \textbf{M1} on `Not Hate').
Note that \textsc{HateCheck} only has negative predictive power. These results indicate the \textit{absence} of particular weaknesses in models rather than necessarily characterising generalisable strengths.

A further caveat is that in \textbf{R2} the annotators were given adversarial pivots to improve their ability to trick the models (See above). These pivots exploit similar model weaknesses as the functional tests in \textsc{HateCheck} expose, which creates a risk that this gold standard is not truly independent. We did not identify any exact matches, although after lowering case and removing punctuation there are 21 matches. This is just 0.05\% of our dataset but indicates a risk of potential overlap and cross-dataset similarity.

\section{Discussion}
Online hate detection is a complex and nuanced problem, and creating systems that are accurate, robust and generalisable across target, type and domain has proven difficult for AI-based solutions. 
It requires having datasets which are large, varied, expertly annotated and contain challenging content.
Dynamic dataset generation offers a powerful and scalable way of creating these datasets, and training and evaluating more robust and high performing models.  
Over the four rounds of model training and evaluation we show that the performance of target models improves, as measured by their accuracy on the test sets. 
The robustness of the target models from later rounds also increases, as shown by their better performance on \textsc{HateCheck}. 

Dynamic data creation systems offer several advantages for training better performing models.
First, problems can be addressed as work is conducted -- rather than creating the dataset and then discovering any inadvertent design flaws.
For instance, we continually worked with annotators to improve their understanding of the guidelines and strategies for tricking the model. We also introduced perturbations to ensure that content was more challenging. 
Second, annotators can input more challenging content because their work is guided by real-time feedback from the target model. 
Discussion sessions showed that annotators responded to the models' feedback in each round, adjusting their content to find better ways to trick it. 
This process of people trying to find ways to circumvent hate speech models such that their content goes undetected is something that happens often in the real world.
Third, dynamic datasets can be constructed to better meet the requirements of machine learning; our dataset is balanced, comprising ${\sim}54$\% hate. It includes hate targeted against a large number of targets, providing variety for the model to learn from, and many entries were constructed to include known challenging content, such as use of slurs and identity referents. 

However, our approach also presents some challenges.
First, it requires substantial infrastructure and resources. This project would not have been possible without the use of an online interface and a backend that can serve up state-of-the-art hate speech detection models with relatively low latency.
Second, it requires substantial domain expertise from dataset creators as well as annotators, such as knowing where to find real-world hate to inspire synthetic entries. This requires a cross-disciplinary team, combining social science with linguistics and machine learning expertise.
Third, evaluating and validating content in a time-constrained dynamic setting can introduce new pressures on the annotation process. The perturbation process also requires additional annotator training, or else might introduce other inadvertent biases.


\section{Conclusion}

We presented a human-and-model-in-the-loop process for training an online hate detection system. It was employed dynamically to collect four rounds of hate speech datasets. The datasets are large and high quality, having been obtained using only expert annotators. They have fine-grained annotations for the type and target of hate, and include perturbations to increase the dataset difficulty. We demonstrated that the models trained on these dynamically generated datasets are much better at the task of hate speech detection, including evaluation on out-of-domain functional test suites.

In future work we aim to expand the size and diversity of the annotator pool for further rounds of dynamic adversarial data collection. We would like to evaluate different models in-the-loop beyond RoBERTa. The datasets also open many new avenues of investigation, including training models on only original entries and evaluating against perturbations (and vice versa) and training multi-label results for type and target of hate. 
Data collection for future rounds is ongoing.


\section*{Impact Statement \& Ethical Considerations} \label{sec: impact}
In the Impact Statement we address relevant ethical considerations that were not explicitly discussed in the main body of the paper.

\paragraph{Data}
The entries in the dataset were created by the annotation team and, where needed, reviewed by the expert annotators. In no cases did annotators enter content that they found on online sites. All entries which were closely inspired by real-world content (e.g., data entered during round 4) had substantial adjustments made to them. As such, the data is synthetic.

\paragraph{Annotator Compensation}
We employed a team of twenty annotators to enter content who worked varying hours on a flexible basis over four months. 
Annotators were compensated at a rate of \pounds16 per hour.
The rate was set 50\% above the local living wage (\pounds10.85), even though all work was completed remotely.
All training time and meetings were paid.

\paragraph{Intended Use}
The approach, dataset and models presented here are intended to support more accurate and robust detection and classification of online hate.
We anticipate that the high-quality and fine-grained labels in the dataset will advance research in online hate in other ways, such as enabling multiclass classification of types and targets of online hate.

\paragraph{Potential Misuse}
The dataset and models we present could in principle be used to train a generative hate speech model. Alternatively, the dataset and models could be used to better understand the limitations of current detection tools and then attack them. For instance, if a malicious actor investigated our models then they could better understand what content tricks content moderation tools and then use this knowledge to avoid their content being flagged on social media platforms.
However, we believe that these outcomes are unlikely.
We do not report any new weaknesses that have not been established in previous research, and the models we present still contain several limitations.
Further, it is unlikely that a malicious actor would be able to train a powerful enough generative model from this dataset (given its size and composition) to affect their activities.
Overall, the scientific and social benefits of the present research arguably outweighs the small risk of their misuse.

\bibliography{anthology, custom}
\bibliographystyle{acl_natbib}

\clearpage
\appendix
\section{\label{sec:appendix-identities}List of identities}
See Table~\ref{tab:listidentities} for a list of identities focused on during data collection. 

\begin{table}[ht]
\begin{tabular}{ll}
                                &                                                        \\
\textbf{Category of identity}   & \textbf{Identity}                                      \\
Disability & \begin{tabular}{@{}l@{}}People with\\disabilities\end{tabular}                              \\
Gender                          & \begin{tabular}{@{}l@{}}Gender minorities\\(e.g. non binary)\end{tabular}                \\
Gender                          & Women                                                  \\
Gender                          & Trans                                                  \\
Immigration status              & Immigrants                                             \\
Immigration status              & Foreigner                                              \\
Immigration status              & Refugee                                                \\
Immigration status              & Asylum seeker                                          \\
Race / Ethnicity                & Black people                                           \\
Race / Ethnicity                & Indigenous                                             \\
Race / Ethnicity                & \begin{tabular}{@{}l@{}}East Asians\\(e.g. China)\end{tabular}                                    \\
Race / Ethnicity                & \begin{tabular}{@{}l@{}}East Asians\\(e.g. Korea)\end{tabular}               \\
Race / Ethnicity                & \begin{tabular}{@{}l@{}}South East Asians \\(e.g. India)\end{tabular}                  \\
Race / Ethnicity                & Pakistanis                                             \\
Race / Ethnicity                & \begin{tabular}{@{}l@{}}Aboriginal people\\\begin{tabular}{@{}l@{}}(e.g. Native \\Americans)\end{tabular} \end{tabular}              \\
Race / Ethnicity                & Mixed race                                             \\
Race / Ethnicity                & Minority groups                                        \\
Race / Ethnicity                & Arabs                                                  \\
Race / Ethnicity                & \begin{tabular}{@{}l@{}}Travellers\\(e.g. Roma)\end{tabular} \\
Race / Ethnicity                & People from Africa                                     \\
Religion or belief              & Muslims                                                \\
Religion or belief              & Jews                                                   \\
Sexual orientation              & Gay                                                    \\
Sexual orientation              & Lesbian                                                \\
Sexual orientation              & Bisexual                                               \\
National origin                 & Polish                                                 \\
Religion or belief              & Hindus                                                 \\
Class                           & Working class                                          \\
Race / Ethnicity                & \begin{tabular}{@{}l@{}}Hispanic\\(e.g. Latinx)\end{tabular} \\
Intersectional       & Black women                                            \\
Intersectional       & Black men                                              \\
Intersectional       & Indigenous women                                       \\
Intersectional      & Asian women                                            \\
Intersectional    & Muslim women                                          
\end{tabular}
\caption{List of high priority identities}
    \label{tab:listidentities}
\end{table}

\section{\label{sec:appendix-pivots}List of pivots in \textbf{R2}}

\subsection*{Hate pivots}
\paragraph{Convoluted statements}
Statements which express hate through a convoluted sentence structure, e.g. `I’ll make sure that X don’t exist anymore because that's what I want to do and I dont care if nobody knows it".

\paragraph{Negation of neutrality in the second clause}
Statements which initially appear to be neutral (or positive) about an identity, but then undermine it with a caveat, e.g. `some people say it’s not fair to call X dumb… but 9 times out of 10 it’s true!'.

\paragraph{Long phrases}
Statements which are long, often involving multiple clauses and sentences. In such cases the hate may use more subtle forms of language and/or rely more heavily on tone, e.g. `Italy is a beautiful country and I love it so much, from its food to its fashion, from its history to its arts, but i am very worried about the current trends which remind me of the 1930s, So let's import X to change up the food, fashion, history and arts?'

\paragraph{Rhetorical questions}
Statements which ask a rhetorical question in order to make a hateful point. Their goal is to express hate rather than to elicit an actual answer, e.g. `Aren’t all X penny pinching, after all?'

\paragraph{Misspelling (swapped characters)}
Words which are intentionally misspelt so that, although the meaning is clear to a human, the model misclassifiers it because the hate is ‘obfuscated’.\footnote{Note we use * for the first vowel in each slur to minimize use of harmful language.} e.g. \textit{n*gga}: `N!gg@', `n!gga', `n1gg4’', \textit{f*g}: `f@g', `f4g', `f@gz', \textit{k*ke}: `k1ke', `k1k3'.

\paragraph{Misspelling (elongation)}
Words which are misspelt so that they are far longer, with many repeated letters, e.g. `I think letting gaaaaaaaay people have rights is just soooooooo awful'.

\paragraph{Misspelling (spaces)}
Words which are misspelt so that there are spaces between letters. Misspelling of terms includes: ‘T R A N N Y S’, ‘J e w s’, ‘N i g g a’.

\paragraph{Synonym (hate expressed with infrequently used terms)}
Statements which contain rarely used terms,, e.g. rather than saying ‘X are scum’ they use the phrase, ‘X are all sour misanthropes’. Equally, such statements might just direct unusual / less common forms of abuse at the targeted identity.
Infrequently used terms to insult a group include: ‘Obnoxious’, ‘Heinous’, ‘Abhorrent’, ‘Unlawful’, ‘Dishonorable’.

\paragraph{Terms with positive sentiment}
Statements which contain positive phrases and terms but still express hate. In particular, phrases which use the term 'love' are often identified as not hateful, e.g. `I just love it when all of the brown people leave my office so I dont have to look at them'.

\paragraph{Random statement}
Statements which are 'random' and do not content any features which are likely to trick the model.

\subsection*{Not Hate pivots}

\paragraph{Counter speech (by another identity)}
Statements which challenge, counter or dismiss hate, which are expressed by someone who does \textit{not} identify with the group being attacked. This often involves referring to or quoting hateful content -- which models may mistake for genuine hate.

\paragraph{Counter speech (by target identity)}
Statements which challenge, counter or dismiss hate, which are expressed by someone who is from the group being attacked.

\paragraph{Counter speech (use of negation)}
Statements which are neutral and state that a group does not have a (negative or neutral) attribute. This is presented primarily as a contrast set to \textit{Negated non-hate} (in the Hateful pivots), e.g. `Jews are not money lovers'.

\paragraph{Not hateful use of a term that is also a referent for an identity (polysemy)}
Statements that mention a referent for an identity (in a way that is not meant to refer to the identity).
For instance, using `black' as a colour, e.g. `the black cat down the road really ruins the neighbourhood'.

\paragraph{Use of profanities}
Statements which contain a profanity in a not hateful way.
They are often used as an adjective or adverb to describe an emotion or to place emphasis on what is being said, e.g. `fucking hell today was a lot of bullshit'.

\paragraph{Negativity against objects}
Statements which attack, criticise or express negativity against inanimate objects, such as sofa or a car, e.g. `this cup is totally rubbish'.

\paragraph{Personal abuse (direct)}
Statements which are aggressive, insulting or abusive against an individual using a direct personal pronoun (i.e. ‘you’), e.g. `you are a complete joke and no-one respects you, loser'.

\paragraph{Personal abuse (indirect)}
Statements which are aggressive, insulting or abusive against an individual who is not part of the conversation and as such is referred to with an indirect personal pronoun (i.e. ‘he’, ‘she’, ‘they’), e.g. `he is such a waste of space. I hope he dies'.

\paragraph{Negativity against concepts}
Statements which attack, criticise or express negativity against concepts and ideologies, such as political ideologies, economic ideas and philosophical ideals, e.g. `I've never trusted capitalism. It's bullshit and it fucks society over'.

\paragraph{Negativity against animals}
Statements which attack, criticise or express negativity against animals, e.g. `dogs are just beasts, kick them if they annoy you'.

\paragraph{Negativity against institutions}
Statements which attack, criticise or express negativity against institutions; such as large organisations, governments and bodies, e.g. `the NHS is a badly run and pointless organisation which is the source of so much harm'.

\paragraph{Negativity against others}
Statements which attack, criticise or express negativity against something that is NOT an identity -- and the targets are not identified elsewhere in this typology, e.g. `the air round here is toxic, it smells like terrible'.

\section{\label{sec:appendix-dev}Development set performance}

Table~\ref{tab:performancedev} shows dev set performance numbers.

\begin{table*}[t]
    \centering
    \begin{tabular}{l|cccc}\toprule
    \textbf{Model} & \textbf{R1} & \textbf{R2} & \textbf{R3} & \textbf{R4} \\\midrule
    M1 (R1 Target) & 41.4$\pm$0.91 & 61.06$\pm$0.43 & 58.18$\pm$0.69 & 55.46$\pm$0.63 \\ 
    M2 (R2 Target)& \textbf{95.38$\pm$0.25} & 68.86$\pm$0.71 & 66.46$\pm$1.09 & 63.17$\pm$0.8 \\ 
    M3 (R3 Target) & 94.55$\pm$0.65 & 85.04$\pm$0.63 & 76.77$\pm$0.57 & 74.4$\pm$0.9 \\ 
    M4 (R4 Target) & 94.92$\pm$0.45 & \textbf{85.32$\pm$0.29} & \underline{\textbf{77.52$\pm$0.68}} & \underline{\textbf{76.42$\pm$0.82}} \\ \midrule
    M(R1) & \underline{\textbf{95.69$\pm$0.29}} & 61.88$\pm$0.98 & 57.75$\pm$0.8 & 58.54$\pm$0.52 \\ 
    M(R2) & 81.28$\pm$0.2 & \textbf{84.36$\pm$0.4} & 75.8$\pm$0.55 & \textbf{74.29$\pm$1.05} \\ 
    M(R3) & 76.79$\pm$1.18 & 79.6$\pm$0.99 & 75.5$\pm$0.48 & 74.19$\pm$1.07 \\ 
    M(R4) & 78.05$\pm$1.09 & 80.21$\pm$0.31 & \textbf{75.63$\pm$0.49} & 72.54$\pm$0.64 \\ \midrule
    M(R0+R1) & \textbf{93.92$\pm$0.3} & 69.43$\pm$1.58 & 65.48$\pm$0.48 & 63.99$\pm$0.74 \\ 
    M(R0+R1+R2) & 93.13$\pm$0.24 & 82.82$\pm$0.8 & 73.66$\pm$0.75 & 72.28$\pm$0.84 \\ 
    M(R0+R1+R2+R3) & 93.43$\pm$0.39 & 84.66$\pm$0.6 & 75.81$\pm$0.29 & \textbf{75.85$\pm$1.0} \\ 
    M(R0+R1+R2+R3+R4) & 92.73$\pm$0.82 & \underline{\textbf{86.0$\pm$0.69}} & \textbf{77.0$\pm$0.59} & 75.7$\pm$0.69 \\\bottomrule
    \end{tabular} 
    \caption{Macro F1 with standard deviation over 5 training rounds, evaluated on each rounds' dev set. Early-stopping is performed on the latest development set for each round where dev results are obtained at least once per epoch, out of four epochs.}
    \label{tab:performancedev}
\end{table*}

\section{\label{sec:appendix-model}Model, Training, and Evaluation Details}

The model architecture was the roberta-base model from Huggingface (\url{https://huggingface.co/}), with a sequence classification head. This model has approximately 125 million parameters. Training each model took no longer than approximately a day, on average, with 8 GPUs on the FAIR cluster. All models were trained with a learning rate of 2e-5 with the default optimizer that Huggingface's sequence classification routine uses. Target model hyperparameter search was as follows: the R2 target was trained for 3 epochs on the R1 target training data, plus multiples of the round 1 data from \{1, 5, 10, 20, 40, 100\} (the best was 5). The R3 target was trained for 3 epochs on the R2 target training data, plus multiples of the round 2 data from \{1, 5, 10, 20, 40, 100\} (the best was 100). The R4 target was trained on the R3 target training data for 4 epochs, plus multiples of the round 3 data from \{1, 5, 10, 20, 40, 100, 200\} (the best was 1); early stopping based on loss on the dev set (measured multiple times per epoch) was performed. 
The dev set we used for tuning target models was the latest dev set we had at each round. We did not perform hyperparameter search on the non-target models, with the exception of training 5 seeds of each and early stopping based on dev set loss throughout 4 training epochs. We recall that model performance typically did not vary by much more than 5\% through our hyperparameter searches.

\section{\label{sec:appendix-ds}Data statement}
Following \newcite{bender-friedman-2018-data} we provide a data statement, which documents the process and provenance of the final dataset.

\paragraph{A. CURATION RATIONALE}
In order to study the potential of dynamically generated datasets for improving online hate detection, we used an online interface to generate a large-scale synthetic dataset of 40,000 entries, collected over 4 rounds, with a `model-in-the-loop' design.
Data was not sampled. Instead a team of trained annotators created synthetic content to enter into the interface. 

\paragraph{B. LANGUAGE VARIETY}
All of the content is in English. We opted for English language due to the available annotation team, and resources and the project leaders' expertise. The system that we developed could, in principle, be applied to other languages.

\paragraph{C. SPEAKER DEMOGRAPHICS}
Due to the synthetic nature of the dataset, the speakers are the same as the annotators. 

\paragraph{D. ANNOTATOR DEMOGRAPHICS}
Annotator demographics are reported in the paper, and are reproduced here for fullness.
Annotation guidelines were created at the start of the project and then updated after each round. Annotations guidelines will be publicly released with the dataset. 
We followed the guidance for protecting and monitoring annotator well-being provided by \newcite{vidgen_challenges_2019}.
$20$ annotators were recruited. Ten were recruited to work for $12$ weeks and ten were recruited for the final four weeks. Annotators completed different amounts of work depending on their availability, which is recorded in the dataset.

All annotators attended a project onboarding session, half day training session, at least one one-to-one session and a daily 'standup' meeting when working. They were given a test assignment and guidelines to review before starting work and received feedback after each round. Annotators could ask the experts questions in real-time over a messaging platform. 

Of the $20$ annotators, $35$\% were male and $65$\%  were female.
$65$\% were $18$-$29$ and $35$\% were $30$-$39$.
$10$\% were educated to high school level, $20$\% to undergraduate, $45$\% to taught masters and $25$\% to research degree~(i.e. PhD).
$70$\% were native English speakers and $30$\% were non-native but fluent.
Respondents had a range of nationalities, including British ($60$\%), as well as Polish, Spanish and Iraqi.
Most annotators identified as ethnically white ($70$\%), followed by Middle Eastern ($20$\%) and two or more ethnicities ($10$\%).
Participants all used social media regularly, including $75$\% who used it more than once per day. All participants had seen other people targeted by online abuse before, and $55$\% had been targeted personally.

\paragraph{E. SPEECH SITUATION}
All data was created from $21$st September $2020$ until $14$th January $2021$.
During the project annotators visited a range of online platforms, with adequate care and supervision from the project leaders, to better understand online hate as it appears online.

\paragraph{F. TEXT CHARACTERISTICS}
The composition of the dataset, including the distribution of the Primary label (Hate and Not) and the type (Derogation, Animosity, Threatening, Support, Dehumanization and None Given) is described in the paper.

\end{document}